%% file: arr2022.tex
\pdfoutput=1
\documentclass[11pt]{article}

\usepackage{acl}
\usepackage{svg}
\usepackage{times}
\usepackage{latexsym}

\usepackage[T1]{fontenc}
\usepackage[utf8]{inputenc}

\usepackage{microtype}

\usepackage{multirow}

\usepackage{subcaption}
\usepackage{graphicx}
\usepackage{multirow}
\usepackage{paralist}
\usepackage{hyperref}
\usepackage{arydshln}
\definecolor{ForestGreen}{RGB}{34,139,34}

\title{BanglaParaphrase: A High-Quality Bangla Paraphrase Dataset}

\author{
Ajwad Akil\thanks{~ These authors contributed equally to this work.} , Najrin Sultana\footnotemark[1] , Abhik Bhattacharjee,  Rifat Shahriyar\\
 [3pt]
Bangladesh University of Engineering and Technology (BUET)\\
[3pt]
\texttt{ajwadakillabib@gmail.com},
\texttt{nazrinshukti@gmail.com},\\
\texttt{abhik@ra.cse.buet.ac.bd}, \texttt{rifat@cse.buet.ac.bd}\\
}

\date{}

\begin{document}
\maketitle

\input{0_abstract}
\input{1_introduction}
\input{2_related}
\input{3_paraphrase_dataset}

\input{4_paraphrase_experiments}
\input{7_conclusion}

\iffalse \section*{Acknowledgments}
The acknowledgments should go immediately before the references. Do not number the acknowledgments section.
Do not include this section when submitting your paper for review.\fi

\bibliographystyle{acl_natbib}
\bibliography{anthology, arr2022}

\clearpage
\input{appendix}

\end{document}

%% file: 0_abstract.tex
\begin{abstract}
In this work, we present BanglaParaphrase, a high-quality synthetic Bangla Paraphrase dataset curated by a novel filtering pipeline. We aim to take a step towards alleviating the low resource status of the Bangla language in the NLP domain through the introduction of BanglaParaphrase, which ensures quality by preserving both semantics and diversity, making it particularly useful to enhance other Bangla datasets. We show a detailed comparative analysis between our dataset and models trained on it with other existing works to establish the viability of our synthetic paraphrase data generation pipeline. We are making the dataset and models publicly available at \url{https://github.com/csebuetnlp/banglaparaphrase} to further the state of Bangla NLP.
\end{abstract}

%% file: 1_introduction.tex
\section{Introduction}
Bangla, despite being the seventh most spoken language by the total number of speakers\footnote{\url{https://w.wiki/Pss}} and fifth most spoken language by native speakers\footnote{\url{https://w.wiki/Psq}} is still considered a low resource language in terms of language processing. \citet{joshi-etal-2020-state} have classified Bangla in the language group that has substantial lackings of efforts for labeled data collection and preparation. This lacking is rampant in terms of high-quality datasets for various natural language tasks, including paraphrase generation.

Paraphrases can be roughly defined as pairs of texts that have similar meanings but may differ structurally. So the task of generating paraphrases given a sentence is to generate sentences with different wordings or/and structures to the original sentences while preserving the meaning. Paraphrasing can be a vital tool to assist language understanding tasks such as question answering \cite{pazzani-engelman-1983-knowledge, dong-etal-2017-learning}, style transfer \cite{krishna-etal-2020-reformulating}, semantic parsing \cite{cao-etal-2020-unsupervised-dual}, and data augmentation tasks \cite{gao-etal-2020-paraphrase}.

Paraphrase generation has been a challenging problem in the natural language processing domain as it has several contrasting elements, such as semantics and structures, that must be ensured to obtain a good paraphrase of a sentence. Syntactically Bangla has a different structure than high-resource languages like English and French. The principal word order of the Bangla language is subject-object-verb (SOV). Still, it also allows free word ordering during sentence formation. The pronoun usage in the Bangla language has various forms, such as "very familiar", "familiar", and "polite forms"\footnote{\url{https://en.wikipedia.org/wiki/Bengali_grammar}}. It is imperative to maintain the coherence of these forms throughout a sentence as well as across the paraphrases in a Bangla paraphrase dataset. Following that thread, we create a Bangla Paraphrase dataset ensuring good quality in terms of semantics and diversity. Since generating datasets by manual intervention is time-consuming, we curate our BanglaParaphrase dataset through a pivoting \cite{zhao-etal-2008-pivot} approach, with additional filtering stages to ensure diversity and semantics. We further study the effects of dataset augmentation on a synthetic dataset using masked language modeling. Finally, we demonstrate the quality of our dataset by training baseline models and through comparative analysis with other Bangla paraphrase datasets and models. In summary:

\begin{compactitem}
    \item We present BanglaParaphrase, a synthetic Bangla Paraphrase dataset ensuring both diversity and semantics.
    \item We introduce a novel filtering mechanism for dataset preparation and evaluation.
\end{compactitem}

%% file: 2_related.tex
\section{Related Work}\label{sec:relatedworks}

% \begin{figure*}[]
%   \centering
%   \includesvg[width=0.98\textwidth]{figures/filter.svg}
%   \caption{Filtering Pipeline}
%   \label{fig:filtering_pipeline}
% \end{figure*}

Paraphrase generation datasets and models are heavily dominated by high-resource languages such as English. But for low-resource languages such as Bangla, this domain is less explored. To our knowledge, only \cite{indicnlg} described the use of IndicBART \cite{dabre2021indicbart} to generate paraphrases using the sequence-to-sequence approach for the Bangla language. One of the most challenging barriers to paraphrasing research for low-resource languages is the shortage of good-quality datasets. Among recent work on low-resource paraphrase datasets, \cite{kanerva-etal-2021-finnish} introduced a comprehensive dataset for the Finnish language. The OpusParcus dataset \cite{creutz2018open} consists of paraphrases for six European languages. For Indic languages such as Tamil, Hindi, Punjabi, and Malayalam, \citet{anand2016shared} introduced a paraphrase detection dataset in a shared task. \citet{tapaco} introduced a paraphrase dataset for 73 languages, where there are only about 1400 sentences in total for the Bangla language, mainly consisting of simple sentences.

%% file: 3_paraphrase_dataset.tex
\section{Paraphrase Dataset Generation and Curation}\label{sec:dataset}
\subsection{Synthetic Dataset Generation}\label{synthetic_dataset_generation}
We started by scraping high-quality representative sentences for the Bangla web domain from the RoarBangla website\footnote{\url{https://roar.media/bangla}} and translated them from Bangla to English using the state-of-the-art translation model developed in \cite{hasan-etal-2020-low} with 5 references. For the generated English sentences, 5 new Bangla translations were generated using beam search. Among these multiple generations, only those (original sentence, back-translated sentence) pairs were chosen as candidate datapoints where the LaBSE~\cite{labse} similarity score for both (original Bangla and back-translated Bangla), as well as (original Bangla and translated English) were greater than 0.7\footnote{We chose 0.7 as the LaBSE semantic similarity threshold following \cite{bhattacharjee-etal-2022-banglabert}}. After this process, there were more than 1.364M sentences with multiple references for each source.

\subsection{Novel Filtering Pipeline}\label{filtering_pipeline}
\begin{table*}[]
\centering
\resizebox{\linewidth}{!}
{
\begin{tabular}{|c|c|c|}
\hline
\textbf{Filter Name} & \textbf{Significance} & \textbf{Filtering Parameters}\\ 
\hline
PINC & Ensure diversity in generated paraphrase & 0.65, 0.76, 0.80\\ 
\hline
BERTScore & Preserve semantic coherence with the source & lower 0.91 - 0.93, upper 0.98\\ 
\hline
N-gram repetition & Reduce n-gram repetition during inference & 2 - 4 grams\\ 
\hline
Punctuation & Prevent generating non-terminating sentences during inference & N/A\\ 
\hline

\end{tabular}
}
\caption{Filtering Scheme }\label{tab:filtering-scheme}
\end{table*}
As mentioned in \cite{pinc}, paraphrases must ensure the fluency, semantic similarity, and diversity. To that end, we make use of different metrics evaluating each of these aspects as \textbf{filters}, in a pipelined fashion.

To ensure diversity, we chose \textbf{PINC} \textit{(Paraphrase In N-gram Changes)} among various diversity measuring metrics such as \cite{pinc, ibleu} as it considers the lexical dissimilarity between the source and the candidates. 
% It computes the percentage of n-grams that appear in the candidate sentence but not in the source sentence.
We name this first filter as \textbf{PINC Score Filter}.  To use this metric for filtering, we determined the optimum threshold value empirically by following a plot\footnote{More details are presented in the Appendix} of the data yield against the PINC score, indicating the amount of data having at least a certain amount of PINC score. We chose the threshold value that maximizes the PINC score with over 63.16\% yield.

Since contextualized token embeddings have been shown to be effective for paraphrase detection \cite{bert}, we use BERTScore \cite{bertscore} to ensure semantic similarity between the source and candidates. After our PINC filter, we experimented with BERTScore, which uses the multilingual BERT model \cite{bert} by default. We also experimented with BanglaBERT \cite{bhattacharjee-etal-2022-banglabert} embeddings and decided to use this as our semantic filter since BanglaBERT is a monolingual model performing exceptionally well on Bangla NLU tasks. We select the threshold similar to the PINC filter by following the corresponding plot, and in all of our experiments, we used F1 measure as the filtering metric. We name this second filter as \textbf{BERTScore Filter}. Through a human evaluation\footnote{More details are presented in the ethical considerations section} of 300 randomly chosen samples, we deduced that pairs having BERTScore (with BanglaBERT embeddings) $\geq$ 0.92 were semantically sound and decided to use this as a starting point to figure out our desired threshold.
We further validated our choice of parameters through model-generated paraphrases, with the models trained on filtered datasets using different parameters (detailed in Section \ref{sec:experimental_setup}).

Initially training on the resultant dataset from the previous two filters, we noticed that some of the predicted paraphrases were growing unnecessarily long by repeating parts during inference. As repeated N-grams within the corpus most likely have been the culprit behind this, attempts to ameliorate the issue were made by introducing our third filter, namely \textbf{N-gram Repetition Filter}, where we tested the target side of our dataset to see if there were any N-gram repeats with a value of $N$ from 1 to 4. We obtained less than 200 sentences on the target side with a 2-gram repetition and decided to use $N=2$ for this filter. Additionally, we removed sentences without terminating punctuation from the corpus to ensure a noise-free dataset before proceeding with the training. We term this last filter as \textbf{Punctuation Filter}. The filters, along with their significance and parameters, have been summarised in Table \ref{tab:filtering-scheme}.

% , and the whole pipeline is demonstrated in Figure \ref{fig:filtering_pipeline}.

\subsection{Evaluation Metrics}\label{evaluation_metrics}
Following the work of \cite{niu2020unsupervised}, we used multiple metrics to evaluate several criteria in our generated paraphrase. For \textbf{quality}, we used sacreBLEU \cite{post-2018-call} and ROUGE-L \cite{lin2004rouge}. We used the multilingual ROUGE scoring implementation introduced by \cite{hasan-etal-2021-xlsum} which supports Bangla stemming and tokenization. For \textbf{syntactic diversity}, we used the PINC score as we did for filtering. For measuring \textbf{semantic correctness}, we used BERTScore F1-measure with BanglaBERT embeddings. Additionally, we used a modified version of a hybrid score named BERT-iBLEU score \cite{niu2020unsupervised} where we also used BanglaBERT embeddings for the BERTScore part.
This hybrid score measures semantic similarity while penalizing syntactical similarity to ensure the diversity of the paraphrases. More details about evaluation scores can be found in the Appendix.

\subsection{Diverse Dataset Generation by Masked Language Modeling}\label{mlm-sub}
We wondered whether the dataset could be further augmented through replacing tokens from a particular part of speech with other synonymous tokens.

To that end, we fine-tuned BanglaBERT ~\cite{bhattacharjee-etal-2022-banglabert} for POS tagging with a token classification head on the \cite{ldcpos1} dataset containing 30 POS tags.

The idea of augmenting the dataset with masking follows the work of \cite{https://doi.org/10.48550/arxiv.2106.05141}. We first tagged the parts of speech of the source side of our synthetic dataset and then chose 7 Bangla parts of speech to maximize the diversification in syntactic content. We masked the corresponding tokens and filled them through MLM sequentially. We used both XLM-RoBERTa~\cite{DBLP:journals/corr/abs-1911-02116} and BanglaBERT to perform MLM out of the box. Of these two, BanglaBERT performed mask-filling with less noise, and thus we selected the results of this model. To ensure consistency with our initial dataset, we also filtered these with our pipeline outlined in Section \ref{filtering_pipeline} by choosing the PINC score threshold of 0.7\footnote{We lowered the threshold since this augmentation does not diversify in terms of the structure of the sentences} and (0.92 - 0.98) (lower and upper limit) for the BERTScore threshold, obtaining about 70K sentences. We used this dataset for training models with our initially filtered one in a separate experiment.\footnote{Further details of the whole experiment can be found in the Appendix.}

%% file: 4_paraphrase_experiments.tex
\begin{table*}[h]
\centering
\resizebox{\linewidth}{!}
{
\begin{tabular}{|c|c|c|c|c|c|c|}
\hline
\textbf{Test Set} & \textbf{Model} & \textbf{sacreBLEU} & \textbf{ROUGE-L} & \textbf{PINC} & \textbf{BERTScore} & \textbf{BERT-iBLEU}\\ 
\hline
\multirow{6}{*}{BanglaParaphrase}& mT5-small & \underline{20.9} & 53.57 & 80.5 & \underline{94.20} & \textbf{92.67}\\

& mT5-small-aug & 19.90 & \underline{53.63} & \underline{80.72} & 94.00 &  \underline{92.54}\\

& BanglaT5 & \textbf{32.8} & \textbf{63.58} & 74.40 & \textbf{94.80} & 92.18\\

& BanglaT5-aug & 32.5 & 63.43 & 74.41 & 94.80 & 92.18\\

& IndicBART & 5.60 & 35.61 & 80.26 & 91.50 & 91.16\\

& IndicBARTSS & 4.90 & 33.66 & \textbf{82.10} & 91.10 & 90.95\\

\hline
\multirow{6}{*}{IndicParaphrase}& mT5-small & 7.3 & 18.66 & \underline{82.30} & \underline{94.30} & \underline{89.06} \\

& mT5-small-aug & 7.0 & 18.27 & \textbf{82.80} &  94.10 &  89.00\\

& BanglaT5 & \underline{11.00} & 19.99 & 74.50 & \textbf{94.80} & 87.738\\

& BanglaT5-aug & 11.00 & 20.10 & 74.43 & 94.80  & 87.540 \\

& IndicBART & \textbf{12.00} & \textbf{21.58} & 76.83 &  93.30 &  \textbf{90.65}\\

& IndicBARTSS & 10.7 & \underline{20.59} & 77.60 &  93.10 &  90.54\\
\hline
\end{tabular}
}

\caption{Test results of different models on BanglaParaphrase and IndicParaphrase Test Set where bold items indicate best results and underlined items indicate the runner up}\label{tab:all-models-test-results}
\end{table*}

\section{Experiments and Results}

\subsection{Experimental Setup}
\label{sec:experimental_setup}
We first filtered the synthetic dataset with our 4-stage filtering mechanisms and then fine-tuned mT5-small model~\cite{xue2020mt5}, keeping the default learning rate as 0.001 for 10 epochs. In each of the experiments, we changed the dataset by keeping the model fixed as our objective was to find the threshold for the first two filters for which the metrics on both the validation and the test set of the individual dataset gave us promising results. We conducted several experiments by varying PINC scores from (0.65, 0.76, 0.80) and BERTScore from (0.91, 0.92, 0.93) and 0.98 (lower and upper limit) by following  respective plots.

The evaluation metrics for each experiment were tracked, and we examined how the thresholds affected the metrics for the test set of the dataset we were experimenting with. We finally chose the effective threshold to be \textbf{0.76} for the PINC score and \textbf{0.92 - 0.98}  (lower and upper limit) for BERTScore such that it provides a good balance between good automated evaluation scores and data amount, and obtained \textbf{466630} parallel paraphrase pairs. We fine-tuned mT5-small, and  BanglaT5 \cite{bhattacharjee2022banglanlg} with the BanglaParaphrase training set as well as with a MLM augmented dataset as mentioned in Section \ref{mlm-sub}. For training, validation, and testing purposes, we randomly split the whole dataset into 80:10:10 ratios. We sampled the MLM dataset twice for the second dataset and added it to our initial training and validation set. After augmentation, the dataset consisted of \textbf{603672} parallel pairs with \textbf{551324} pairs for training and \textbf{29016} for validation. We used the same testing set consisting of \textbf{23332} parallel pairs for all the models.\footnote{MLM augmented dataset is for experimental purpose only} And finally we used the IndicBART and IndicBARTSS \cite{dabre2021indicbart} fine-tuned on the IndicParaphrase dataset \cite{indicnlg} to generate predictions and compute the evaluation scores for comparative analysis.

\paragraph{Hyperparameter Tuning} We fine-tuned mT5-small for 10-15 epochs, tuning the learning rate from {3e-4 to 1e-3}. BanglaT5 was fine-tuned for 10 epochs with a learning rate of 5e-4 and a warmup ratio of 0.1. We chose the final models based on the validation performance of the sacreBLEU score. During inference for the mT5-small model, we used top-K \cite{fan-etal-2018-hierarchical} sampling with a value of 50 in combination with top-P sampling with a value of 0.95 along with beam search for generating multiple inferences, which we filter by PINC score of 0.74 followed by max BERTScore. For BanglaT5, the inference was simply made with a beam search with a beam length of 5.

\subsection{Results and Comparison}
In Table \ref{tab:all-models-test-results}, we show how our trained models namely mT5-small, mT5-small-aug\footnote{\textbf{aug} means the models were trained with MLM augmented BanglaParaphrase training set}, BanglaT5 and BanglaT5-aug models as well as IndicBART and IndicBARTSS perform on our released test set and Indic test Set (only Bangla) from IndicParaphrase dataset. A few examples of how mT5-small performs on the BanglaParaphrase test set and a detailed comparison of the IndicParaphrase dataset with our dataset in terms of diversity and semantics can be found in the Appendix. 

For the BanglaParaphrase test set, we observe that all the evaluation scores are almost similar for both mT5-small and BanglaT5 trained on the original dataset as well as the MLM augmented dataset We find that the BanglaT5 model performs best on sacreBLEU, ROUGE-L, and BERTScore for our test set. We also observe that both the IndicBART models achieve lower scores in all the metrics except PINC, which is not sufficient enough to ensure the quality of generated paraphrases. The scores on sacreBLEU and ROUGE-L are particularly low compared to what our trained models achieved. As for the PINC score, IndicBARTSS achieved the highest value, with mT5 models slightly trailing behind. Since all other scores are lower, this high PINC score has low significance. As for the hybrid score, we find that mT5-small trained on the BanglaParaphrase training set achieves the best result on our test set, with BanglaT5 models trailing slightly lower and IndicBART models having a much lower value.

For the IndicParaphrase test set, we observe that mT5 models perform poorly in sacreBLEU and ROUGE-L scores, whereas BanglaT5 models perform very competitively with IndicBART models inspite of being only fine-tuned on our dataset, which has virtually no overlap with IndicParaphrase training set. We also observe that both mT5 and BanglaT5 trained on the BanglaParaphrase training set and augmented training set have similar performance on all the metrics for this test set. We find both the BanglaT5 models achieve the highest BERTScore, beating IndicBART and IndicBARTSS, and both mT5 models trail closely to BanglaT5. So BanglaT5 can generalize well on other datasets. As for the PINC score, we see that mT5-small-aug achieves the highest score among all the models. And finally, for the hybrid score, we find both IndicBART models achieving the best score. We believe the reason for IndicBART to have higher scores is that it has a high PINC score, i.e., less similarity with the source, which results in a higher BERT-iBLEU score.  

Overall, the models trained on the BanglaParaphrase data set, specifically BanglaT5, perform competitively with the IndicBART models, even besting in terms of semantics concerning the source, while generating diverse paraphrases and thus validating that our dataset not only ensures good diversity but semantics as well.

%% file: 7_conclusion.tex
\section{Conclusion \& Future Works}
In this work, starting from a pure synthetic paraphrase dataset, we introduced an automated  filtering pipeline to curate a high-quality Bangla Paraphrase dataset, ensuring both diversity and semantics. We trained the mT5-small and BanglaT5 models with our dataset to generate quality paraphrases of Bangla sentences. Our choice of the initial monolingual corpus has been made to include highly representative sentences for the Bangla language, which is large enough for an isolated paraphrase generation task. The corpus can easily be extended for desired pretraining tasks using a larger monolingual corpus. Furthermore, we plan on improving the MLM scheme by automating parts of speech selection and using LaBSE with BanglaBERT embeddings to compare semantics at the sentence level, which would ensure better filters and better evaluation of generated paraphrases. Though our work is language-agnostic, the extent to which our approach applies to other low-resource languages given language-specific components (datasets and models) is subject to further experimentation. In future work, we want to investigate the viability of our synthetic data generation pipeline in the context of paraphrase datasets in different languages included in popular benchmarks such as \cite{Gehrmann2022GEMv2MN}. Additionally, we want to investigate how our paraphrase dataset and models can be used to improve the performance of other low-resource tasks in Bangla, such as Readability detection \cite{Chakraborty_Nayeem_Ahmad_2021} and Cross-lingual summarization \cite{bhattacharjee2022crosssum}

\section*{Acknowledgements}
We would like to thank the Research and Innovation Centre for Science and Engineering (RISE), BUET, for funding the project. 

\section*{Ethical Considerations}
\paragraph{Dataset and Model Release} The \emph{Copy Right Act, 2000}\footnote{\url{http://bdlaws.minlaw.gov.bd/act-details-846.html}} of Bangladesh allows public release and reproduction and  of copy-right materials for non-commercial research purposes. As valuable research work for Bangla Language, we will release the BanglaParaphrase dataset under a non-commercial license. Additionally, we will release the relevant codes and the trained models for which we know the distribution will not cause copyright infringement.

\paragraph{Manual Efforts} 
% \medskip
% \noindent\textbf{Quality Control in Human Translation} 
The manual observations regarding the choice of primary BERTScore threshold which is reflective of high semantic quality by going through 300 randomly chosen samples were done by the native authors.

%% file: appendix.tex
\section*{Appendix}

\subsection*{PINC Score Details}
PINC score is defined as for source sentence
$s$ and candidate sentence $c$ as:

{\large\[
\frac{1}{N}\sum_{n=1}^N 1 - \frac{\mid ngram_s \cap ngram_c \mid}{\mid ngram_c \mid}
\]}

\begin{figure}[h]
  \centering
  \includegraphics[width=0.52\textwidth]{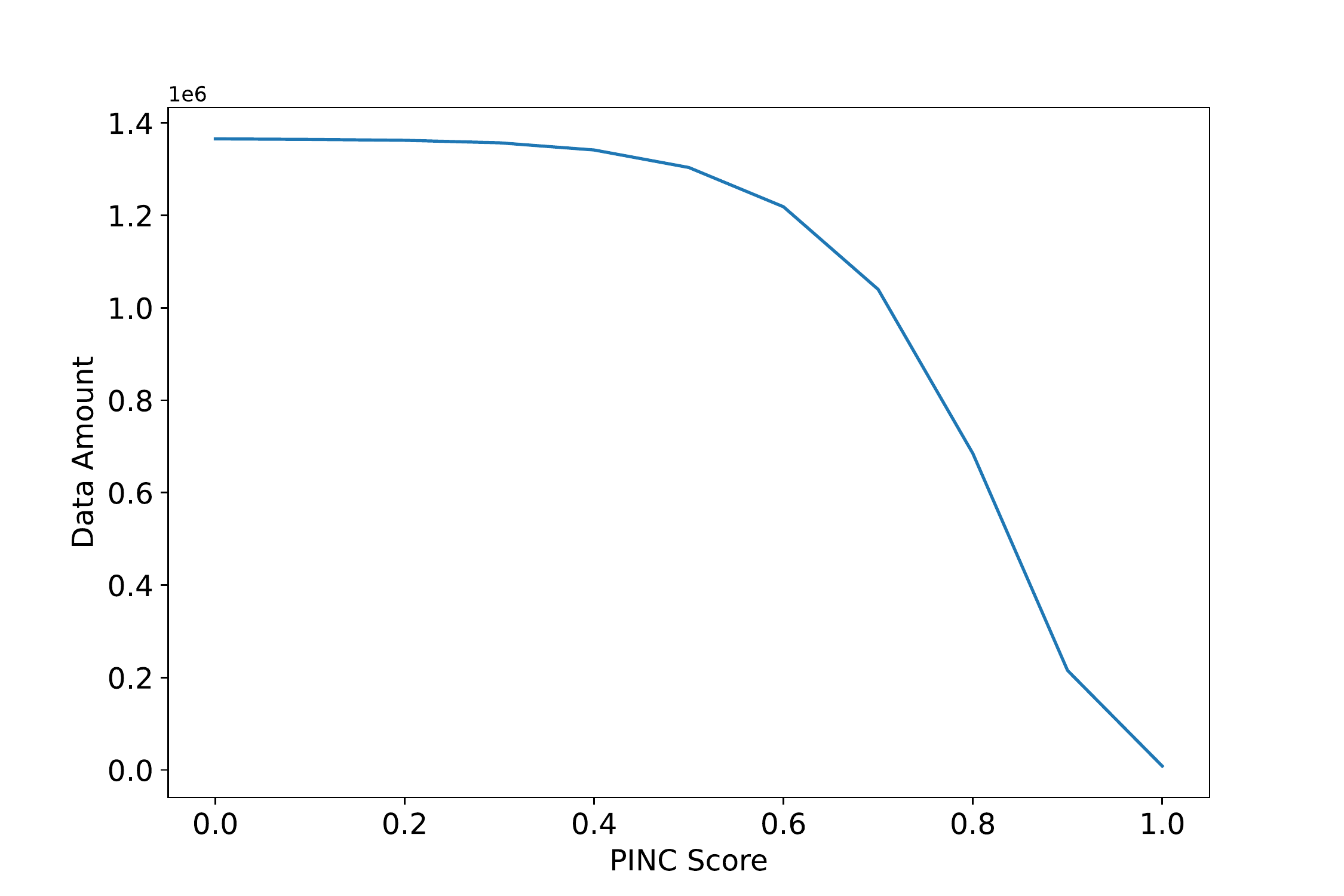}
	\caption{PINC Score range within [0-1] for whole BanglaParaphrase dataset}
	\label{fig:pinc_whole_0_1}
\end{figure}

Where $N$ is defined as the maximum n-gram we considered, and $ngram_s$ and $ngram_c$ are the lists of n-grams present in the source and candidate sentences. In all experiments, we use $N = 4$. This score can be treated as the inverse of the BLEU score since it minimizes the number of n-gram overlaps between the two sentences. We also present a PINC score vs. data amount plot in Figure \ref{fig:pinc_whole_0_1}, which we used to select the thresholds.

\subsection*{BERTScore Plot}

\begin{figure}[h]
  \centering
  \includegraphics[width=0.52\textwidth]{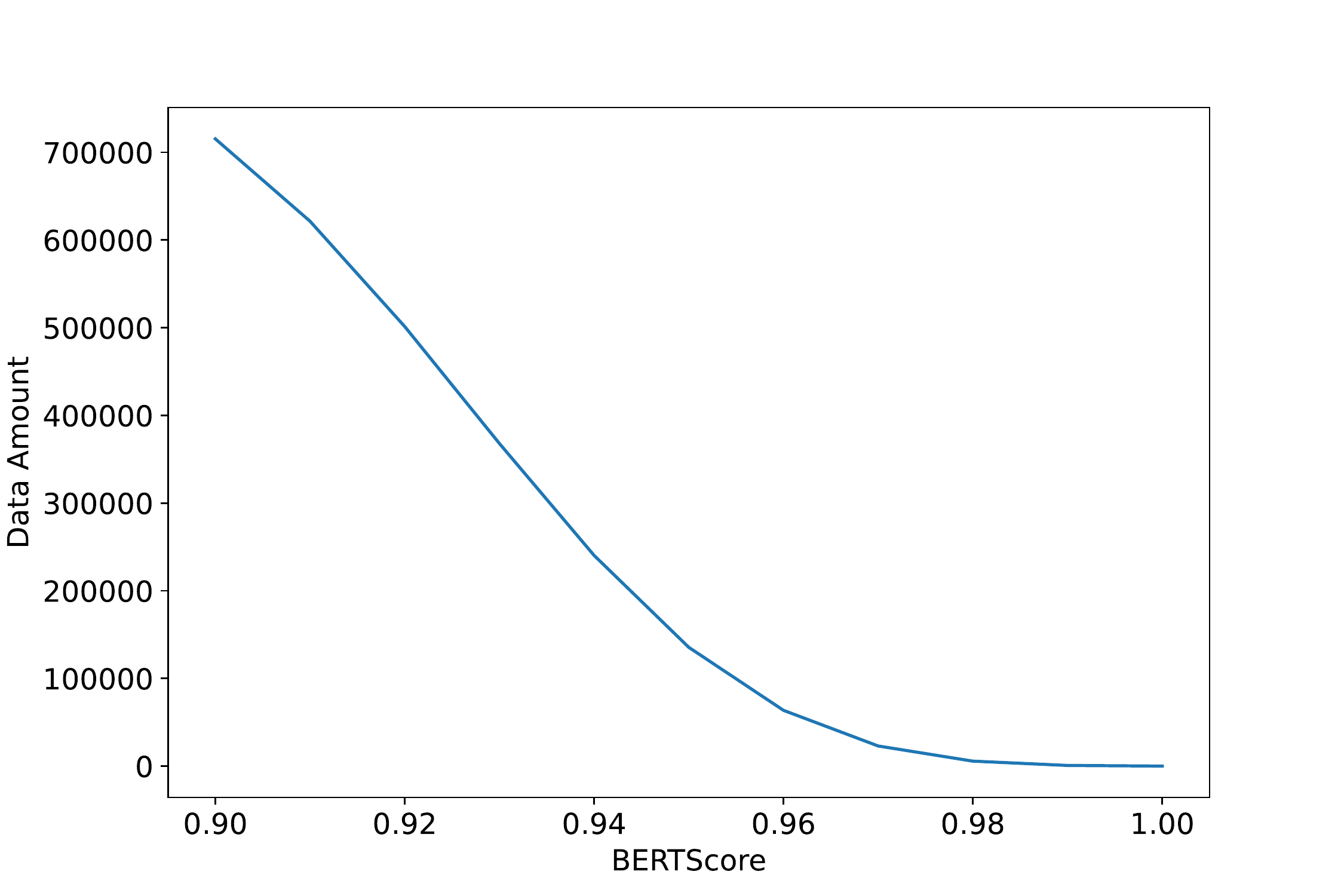}
	\caption{BERTScore with BanglaBERT embeddings within range [0.9-1.0] after whole dataset being filtered by PINC threshold of 0.76}
	\label{fig:BBERTScore_0.9_1.0}
\end{figure}

A plot of BERTScore with BanglaBERT embeddings after the BanglaParaphrase dataset has been filtered with a PINC score of 0.76 threshold is shown in Figure \ref{fig:BBERTScore_0.9_1.0}.

\subsection*{Evaluation Metric Details}
BLEU, METEOR, and ROUGE-L are the most common metrics used \cite{zhou-bhat-2021-paraphrase} for paraphrase evaluation. 
% A generated paraphrase which may differ from the target can still be of good quality. Besides, it may have a high bleu score despite being poor in terms of quality. 
BLEU \cite{bleu} is a widely used metric for machine translation evaluation that ensures semantic adequacy and fluency. But it falls short for paraphrase evaluation as mentioned by \cite{niu2020unsupervised, zhou-bhat-2021-paraphrase}. A unified metric that captures all the elements of evaluating paraphrase is still lacking \cite{zhou-bhat-2021-paraphrase}, and so we present the details about different evaluation metrics we used and the criteria they measure:

\paragraph{Quality} To ensure the quality of the generated paraphrases with respect to the target, we used sacreBLEU Score \cite{post-2018-call} and ROUGE-L \cite{lin2004rouge} F1-measure. Both of the scores produce a real number between the range $[0-1]$, and we present the scores in percentages for our results.
    
\paragraph{Syntactic Diversity} To evaluate the diversity between the generated paraphrases and the sources, we used the PINC score \cite{pinc}. This score produces a real number between the range $[0-1]$ and we report the arithmetic mean for all the sentences in the test set and present in terms of percentages for our results.

\paragraph{Semantic Correctness} To evaluate semantic correctness, the arithmetic mean of BERTScore \cite{bertscore} F1-measure between source and predictions is used. As discussed, this is a modified version of BERTScore which uses BanglaBERT embeddings to produce a real number between $[0-1]$, and we present it in terms of percentages for our results.
    
\paragraph{Hybrid Score} And finally, we used a modified version of a hybrid score named BERT-iBLEU introduced in \cite{niu2020unsupervised}. The formula to compute the score is:
    
{\small\[
\left( \frac{\beta * BERTScore^{-1} + 1.0 * (1-selfBLEU)^{-1}}{\beta + 1.0} \right)^{-1}
\]}

This metric measures semantic similarity while penalizing syntactical similarity at the same time. For the semantic similarity part, the authors used BERTScore between target and predictions, which we modified to use BERTScore with BanglaBERT embeddings. For diversity, self-BLEU was calculated between the source and the prediction. The more dissimilar the source is to the candidate, the higher will be the value of 1-selfBLEU. The final score is a weighted harmonic mean between these two scores. We used the value of $\beta$ to be 4.0, as chosen by the authors. The score produces a real number between the range $[0-1]$, and as our modified BERTScore gives us scores in a high range ($>0.9$), the scores produced by this metric is also in high range. We present the score in terms of percentages for our results.

\subsection*{Diverse Dataset Generation Experiment Details}
We trained BanglaBERT with a token classification head with \cite{ldcpos1} dataset containing 30 POS tags and the entire corpus consists of 7393 sentences corresponding to 102937 tokens. We trained for 20 epochs, with a batch size of 32 and a learning rate of 0.00002 with a linear learning rate scheduler. The dataset was split into an 80:10:10 ratio into a train, test, and validation sets. We obtained close to 90\% F1-Score on the test set. The test set metrics are showed in Table \ref{tab:pos_tagging_exp_details}.

\begin{table}[h]
\centering
\resizebox{\linewidth}{!}
{
\begin{tabular}{|c|c|c|c|c|}
\hline
\textbf{Dataset} & \textbf{Accuracy} & \textbf{Precision} & \textbf{Recall} & \textbf{F1-score}\\ 
\hline
Test & 0.924 & 0.896 & 0.900 & 0.898\\ 
\hline

\end{tabular}
}
\caption{Validation and Test metrics for POS tagging experiment}\label{tab:pos_tagging_exp_details}
\vspace{1em}
\end{table}

\begin{figure}[!tbh]
  \centering
  \includegraphics[width=0.48\textwidth]{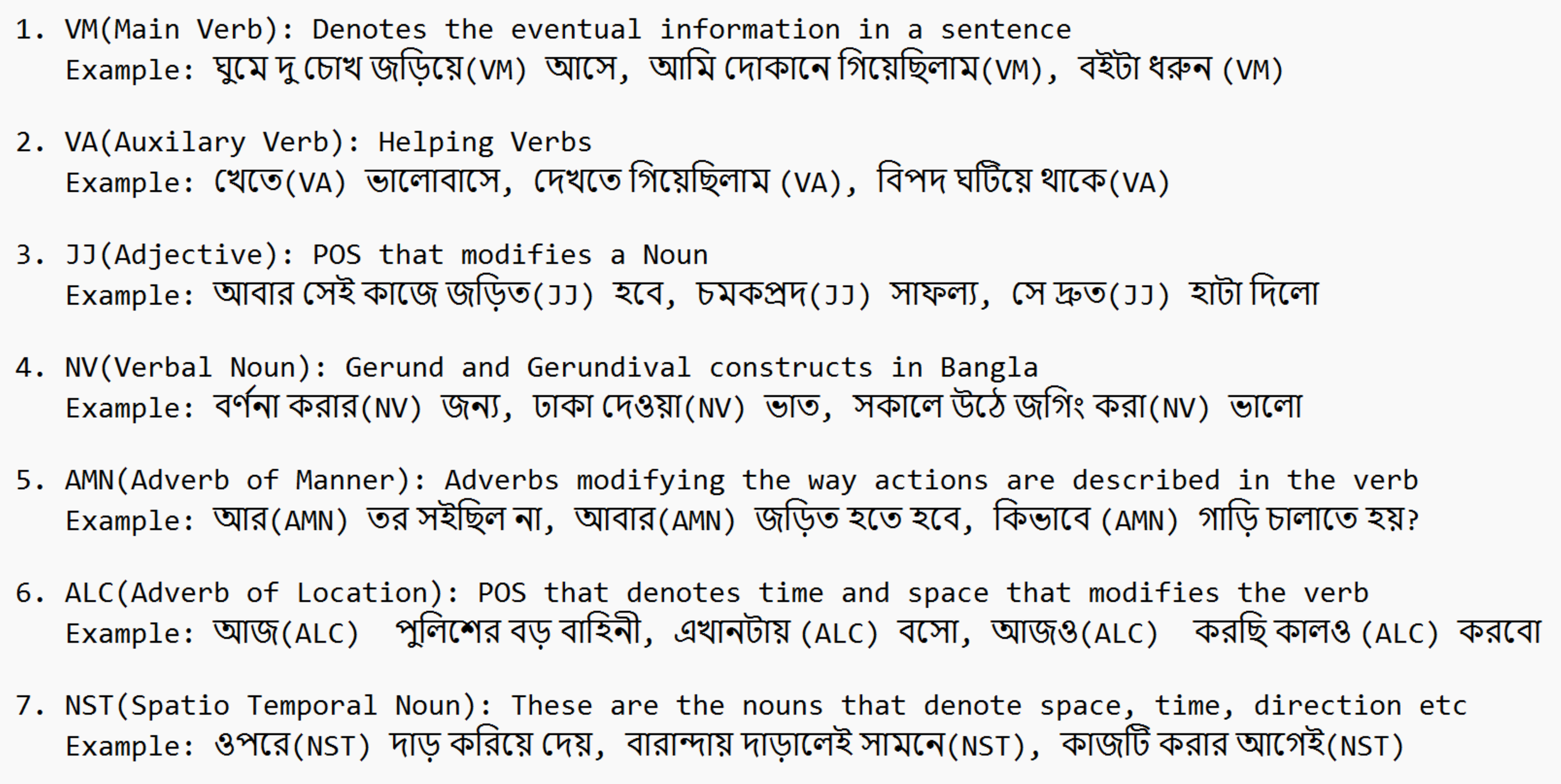}
  \caption{Selected POS Details}
  \label{fig:pos_details}
\end{figure}

\begin{figure}[h]
  \centering
  \includegraphics[width=0.48\textwidth]{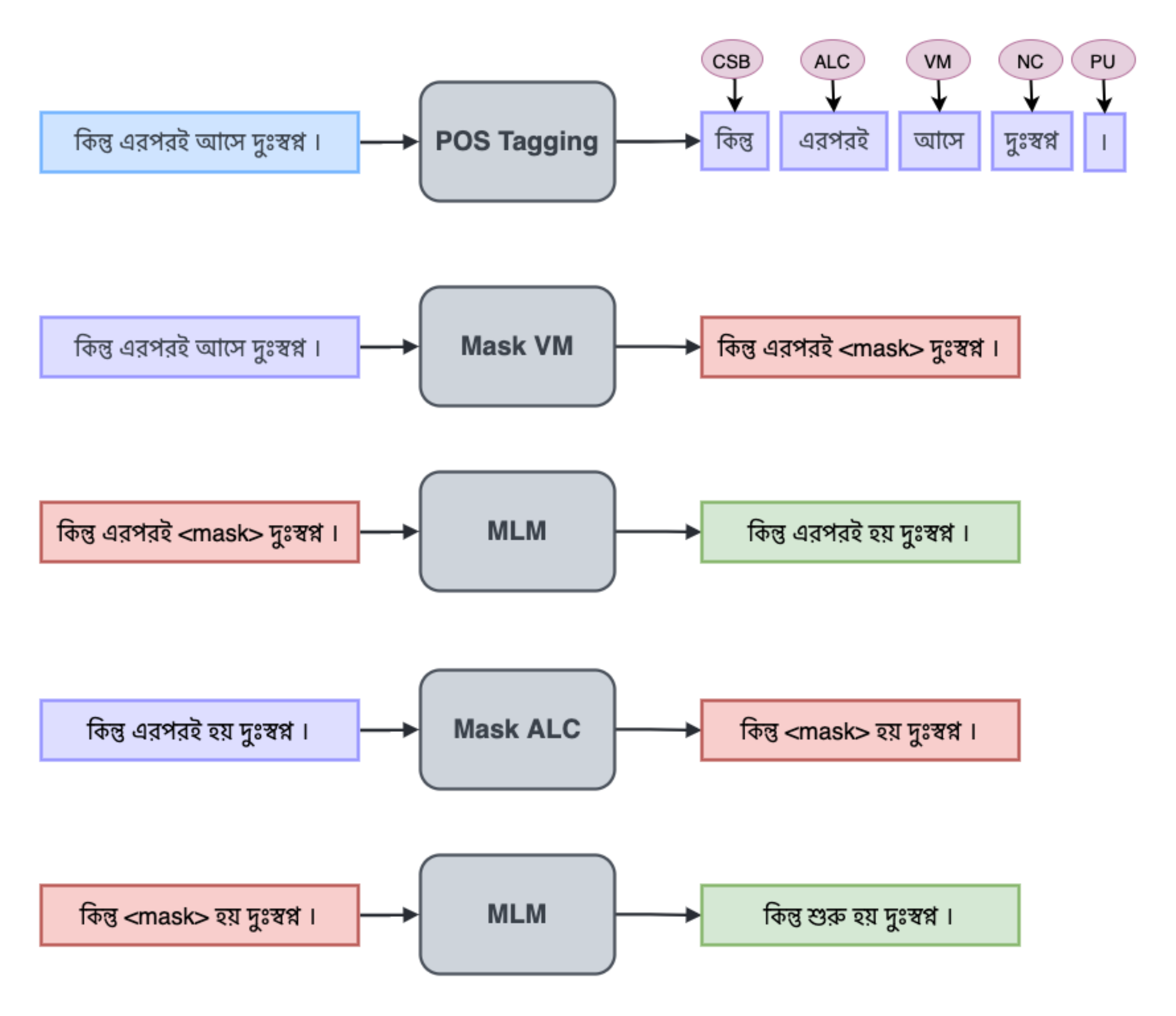}
  \caption{Diverse Sentence Generation by Mask Filling}
  \label{fig:masking_scheme}
\end{figure}

After training the POS tagger, we tagged 7 carefully chosen parts of speeches namely VM (Main verb), VA (Auxilary Verb), JJ (Adjective), NV (Verbal Noun), AMN (Adverb of Manner), ALC (Adverb of location), and NST(Spatio Temporal Noun). These POS were masked and filled in the order as mentioned here. The parts of speeches with minimal description are shown in Figure \ref{fig:pos_details}. A demonstration for mask filling is shown in Figure \ref{fig:masking_scheme}.

\subsection*{Examples of Generated Paraphrase}
\begin{figure}[!tbh]
  \centering
  \includegraphics[width=0.48\textwidth]{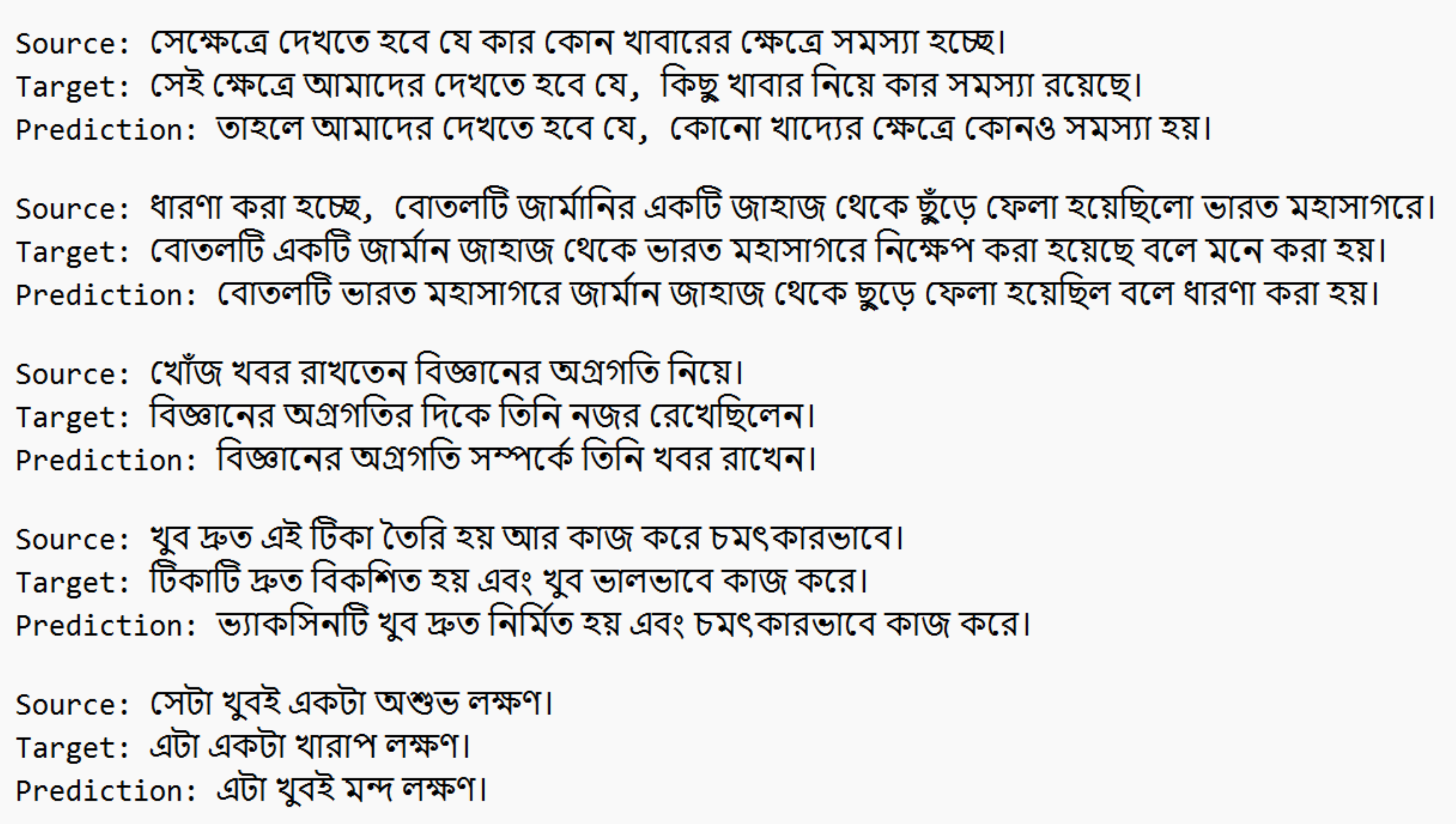}
  \caption{Examples of Generated Paraphrase by mT5 small on released test set (trained with released training set)}
  \label{fig:paraphrase examples}
\end{figure}

We show some examples of generated paraphrases by mT5 small model on BanglaParaphrase dataset in Figure \ref{fig:paraphrase examples}.

% \begin{figure}[!tbh]
%  \centering
%  \includegraphics[width=0.48\textwidth]{figures/sample_complexity_ner.pdf}
%  \vspace{0.0mm}
%  \includegraphics[width=0.48\textwidth]{figures/sample_complexity_qa.pdf}
%  \caption{Sample-efficiency tests with NER and QA.}
%  \label{fig:sample_complexity_qa}
% \end{figure}

\subsection*{BERTScore Distribution Analysis}

\begin{figure}[!tbh]
	\centering
	\begin{subfigure}{0.52\textwidth}
		\includegraphics[width=\textwidth]{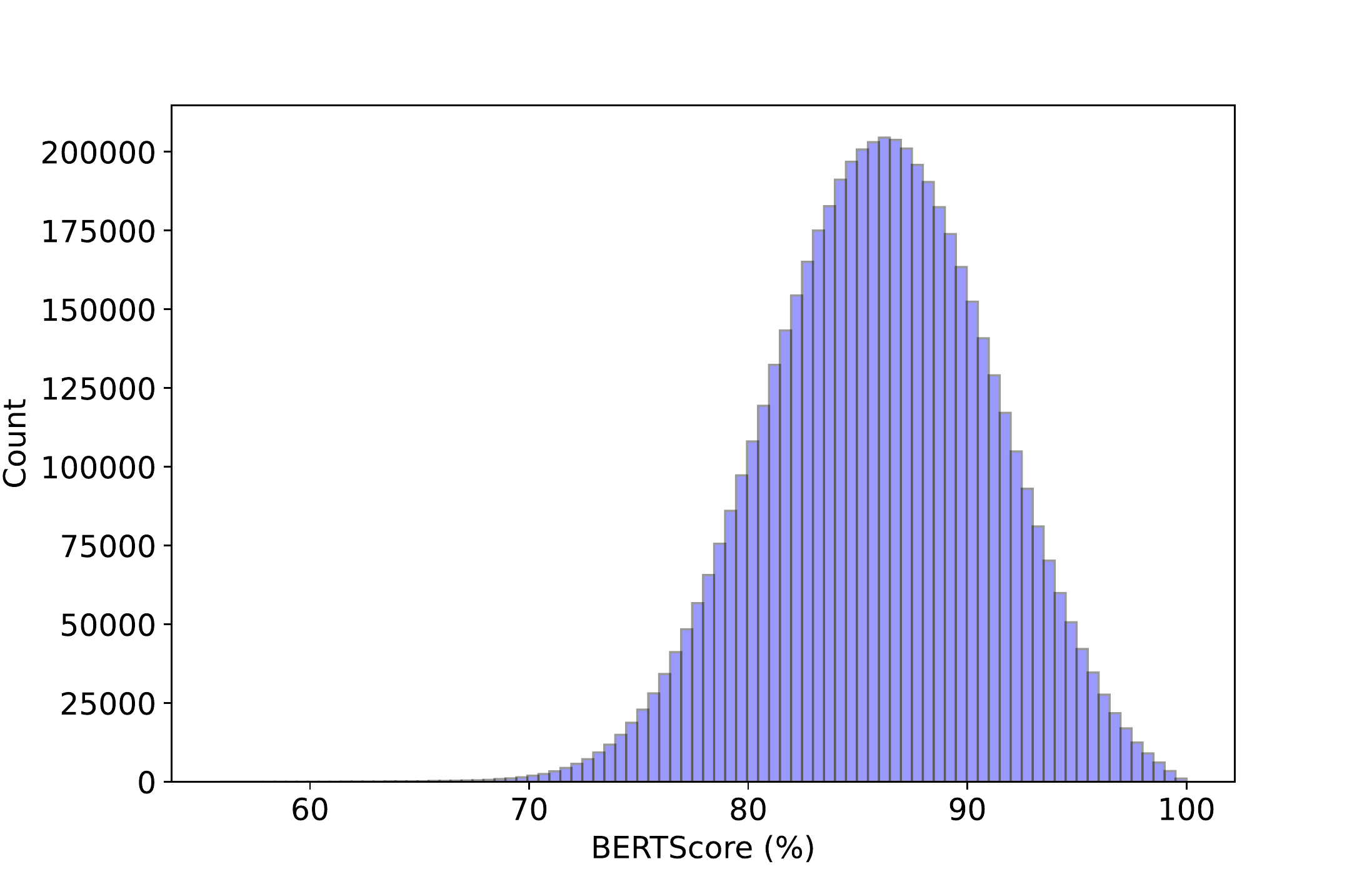}
		\caption{BERTScore Histogram}
		\label{fig:bertscore_histograms}
	\end{subfigure}
	\quad %will generate gap between pictures
	\begin{subfigure}{0.52\textwidth}
		\includegraphics[width=\textwidth]{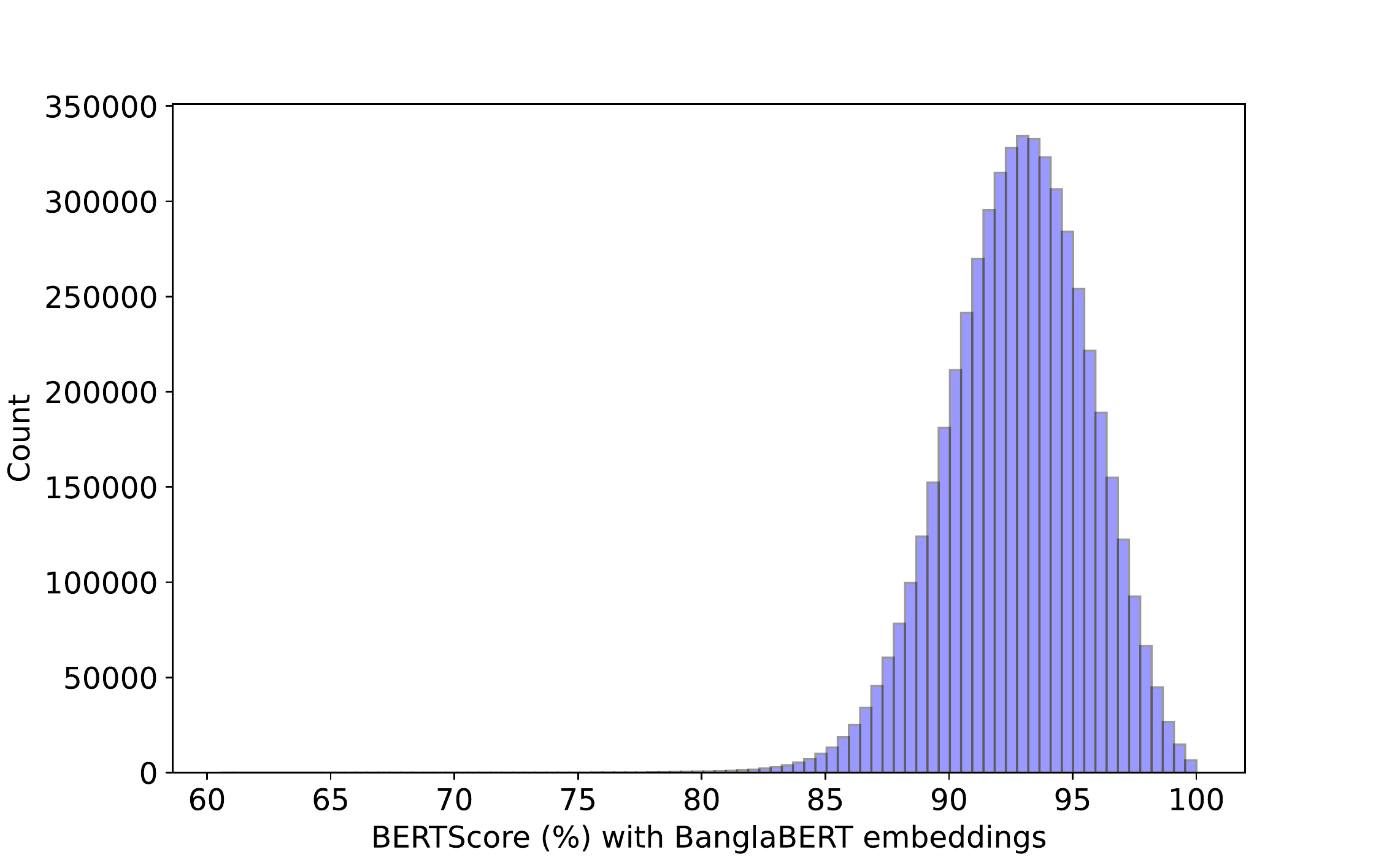}
		\caption{BERTScore Histogram (BanglaBERT embeddings)}
		\label{fig:bbertscore_histograms}
	\end{subfigure}	\caption{Histograms for original dataset}
	\label{fig:histograms}
\end{figure}

BERTScore with mBERT gives us a value in a much more comprehensive range, $[0.7-0.1]$, and most scores are centered around $[0.8-0.9]$ as we can see from the histogram in Figure \ref{fig:bertscore_histograms} whereas BERTScore with BanglaBERT embeddings gives us a score in a much higher range, $[0.8-0.1]$ and most of the scores are centered around $[0.9-0.95]$ as seen in Figure \ref{fig:bbertscore_histograms}. So BERTScore with BanglaBERT embeddings score above 0.8 for sentences with lesser semantic similarity but above 0.9 for sentences with good semantic similarity.

% Similar results are also observed here for the NER task, where BanglaBERT is more sample-efficient when we have $\leq 1k$ training samples. In the QA task, however, both models have identical performance for all sample counts.

% \subsection*{LaBSE similarity analysis for IndicNLG Paraphrasing Dataset}

% \begin{figure}[!tbh]
% 	\centering
% 	\begin{subfigure}{0.48\textwidth}
% 		\includegraphics[width=\textwidth]{figures/indicparaphrasedataset/Indic_paraphrase_LABSE_Plot_all_0.0_1.0.png}
% 		\caption{LaBSE Similarity Score for range [0-1.0]}
% 		\label{fig:labse_indic_all_0_1.0}
% 	\end{subfigure}
% 	\caption{LaBSE Similarity Score for IndicParaphrase dataset}
% 	\label{fig:labse_indic_all}
% \end{figure}

\subsection*{Comparison with IndicNLG Paraphrasing Dataset }
The IndicNLG Suite \cite{indicnlg} has data for eleven languages: 
Assamese, Bangla, Gujarati, Hindi, Marathi, Odiya, Punjabi, Kannada, Malayalam, Tamil, and Telugu. The dataset has 5.57M in size overall. For Bangla Paraphrase, there are 890,445 sentences in the train set, 10,000 in the validation set, and 10,000 in the test set, with each source sentence having 5 references. The dataset uses Samanantar corpus \cite{ramesh2021samanantar} to generate the paraphrases by a back-translation mechanism. Then the authors filtered the sentences by removing noise and duplicates and evaluated the diversity by a scheme developed by them. They screened the sentences in a way to ensure enough diversity among the source and the references. They reported 5 references for each source sentence, which are ordered from most to least diverse. The dataset ensures diversity by a filtering mechanism developed by the authors, but they did not include any filtering mechanism to ensure semantic similarity between the sources or the references. As the initial set of sources and the references were generated by pivoting, there are a lot of
changes and variations and thus, it is vital to ensure both diversity and meaning. 

\begin{figure}[!tbh]
	\centering
	\begin{subfigure}{0.52\textwidth}
		\includegraphics[width=\textwidth]{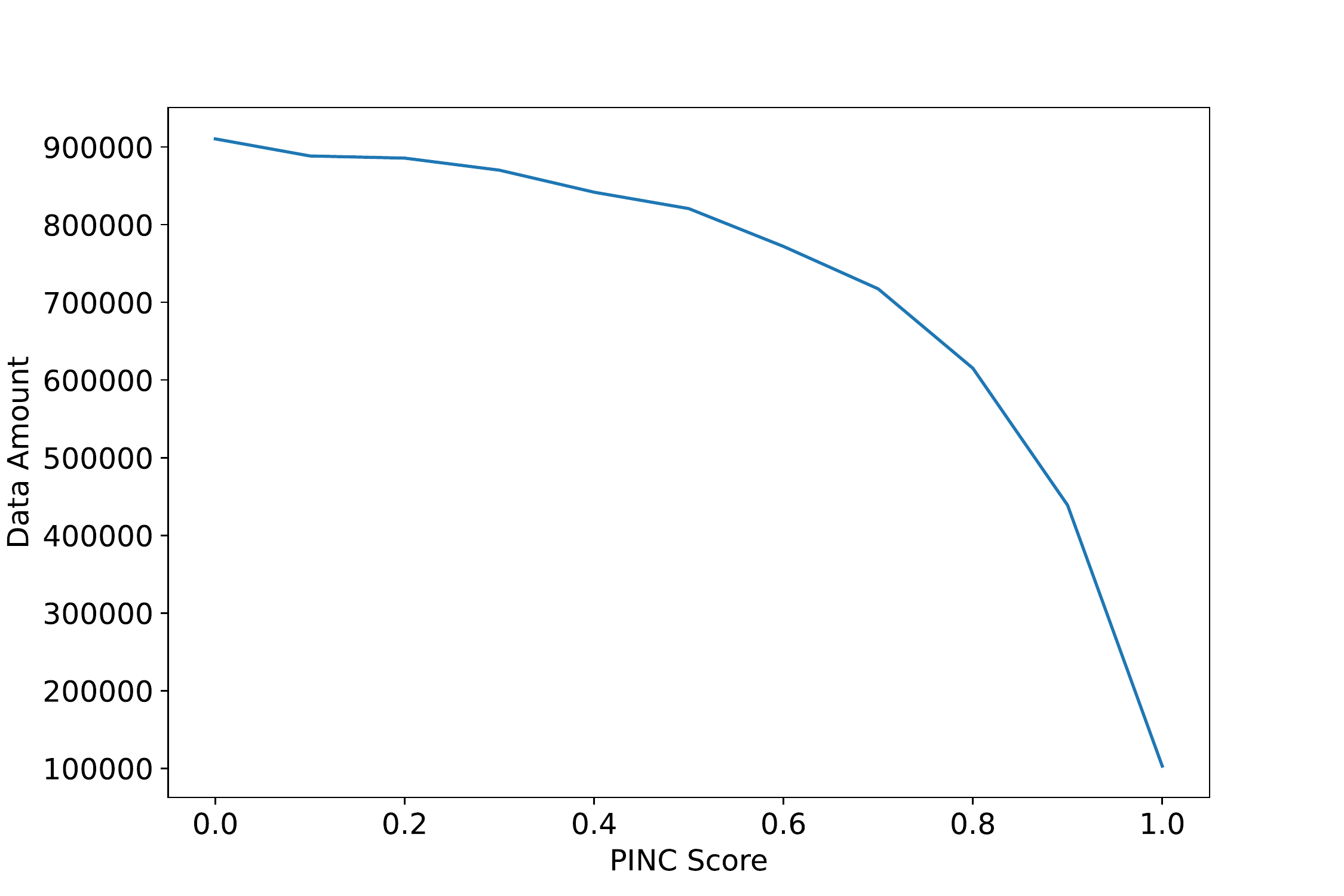}
		\caption{PINC Score for range [0-1.0]}
		\label{fig:pinc_indic_all_0_1.0}
	\end{subfigure}
	\quad %will generate gap between pictures
	\begin{subfigure}{0.52\textwidth}
		\includegraphics[width=\textwidth]{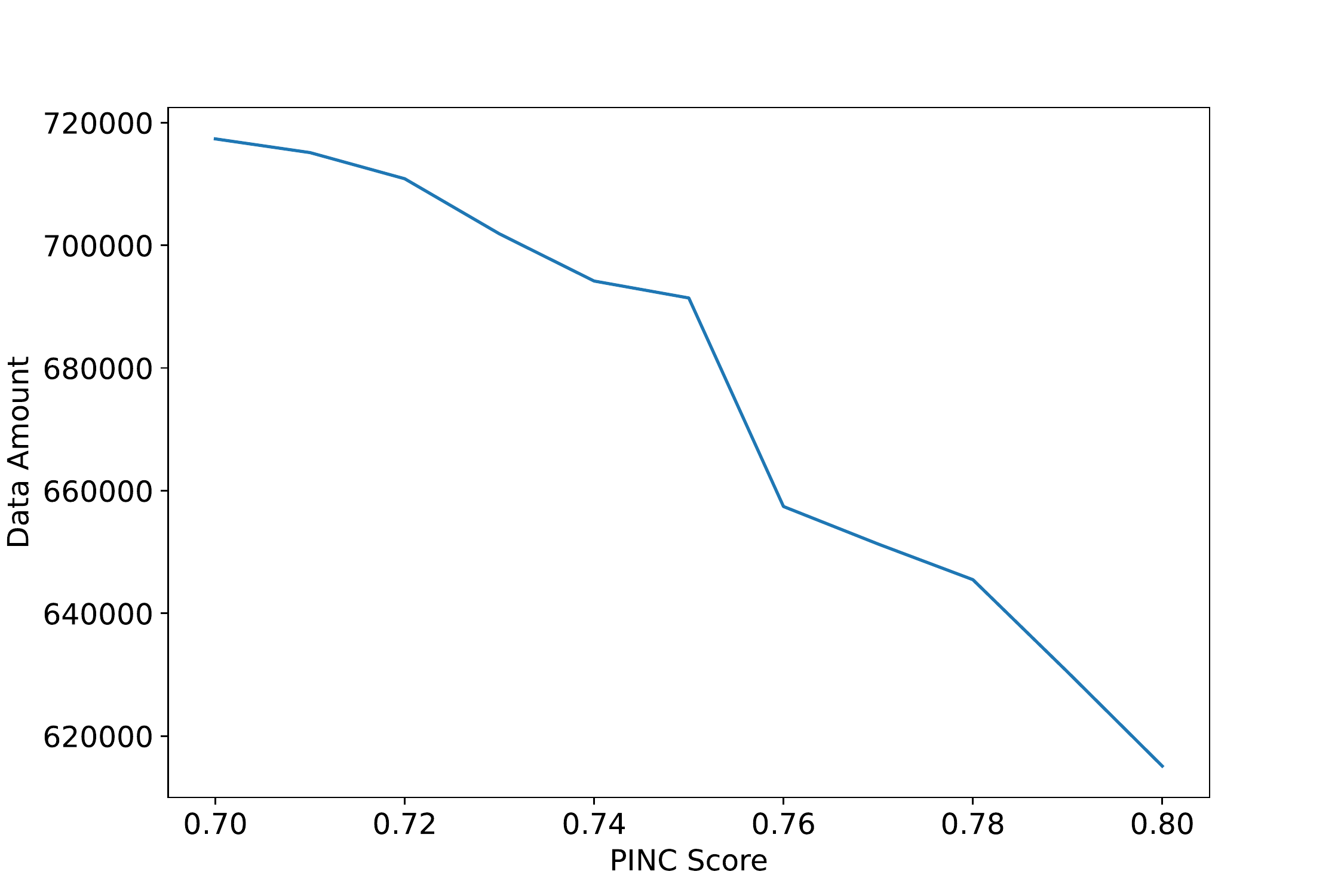}
		\caption{PINC Score for range [0.7-0.8]}
		\label{fig:pinc_indic_all_0.7_0.8}
	\end{subfigure}
	\caption{PINC Score for IndicParaphrase dataset}
	\label{fig:pinc_indic_all}
\end{figure}

To analyze, we plot the scores for the reference with most diversity in terms of PINC score. We started with the PINC score vs. data amount plot in Figure \ref{fig:pinc_indic_all_0_1.0}. The shape of the plot looks a lot similar to the PINC plot for our whole dataset in Figure \ref{fig:pinc_whole_0_1}. We also observe that above or equal to the 0.7 threshold, there are about 0.72M sentences. And for thresholds 0.74 and 0.76, there are about close to 0.7M sentences (about 77\% of the total sentences) and close to 0.66M sentences (about 73\% of the total sentences), respectively. Compared to our filtering, where we chose the PINC filter to be 0.76 and ended up with about 0.86M sentences (about 63.05\% of our total corpus size), the dataset ensured more diverse paraphrases.

\begin{figure}[!tbh]
	\centering
	\includegraphics[width=0.52\textwidth]{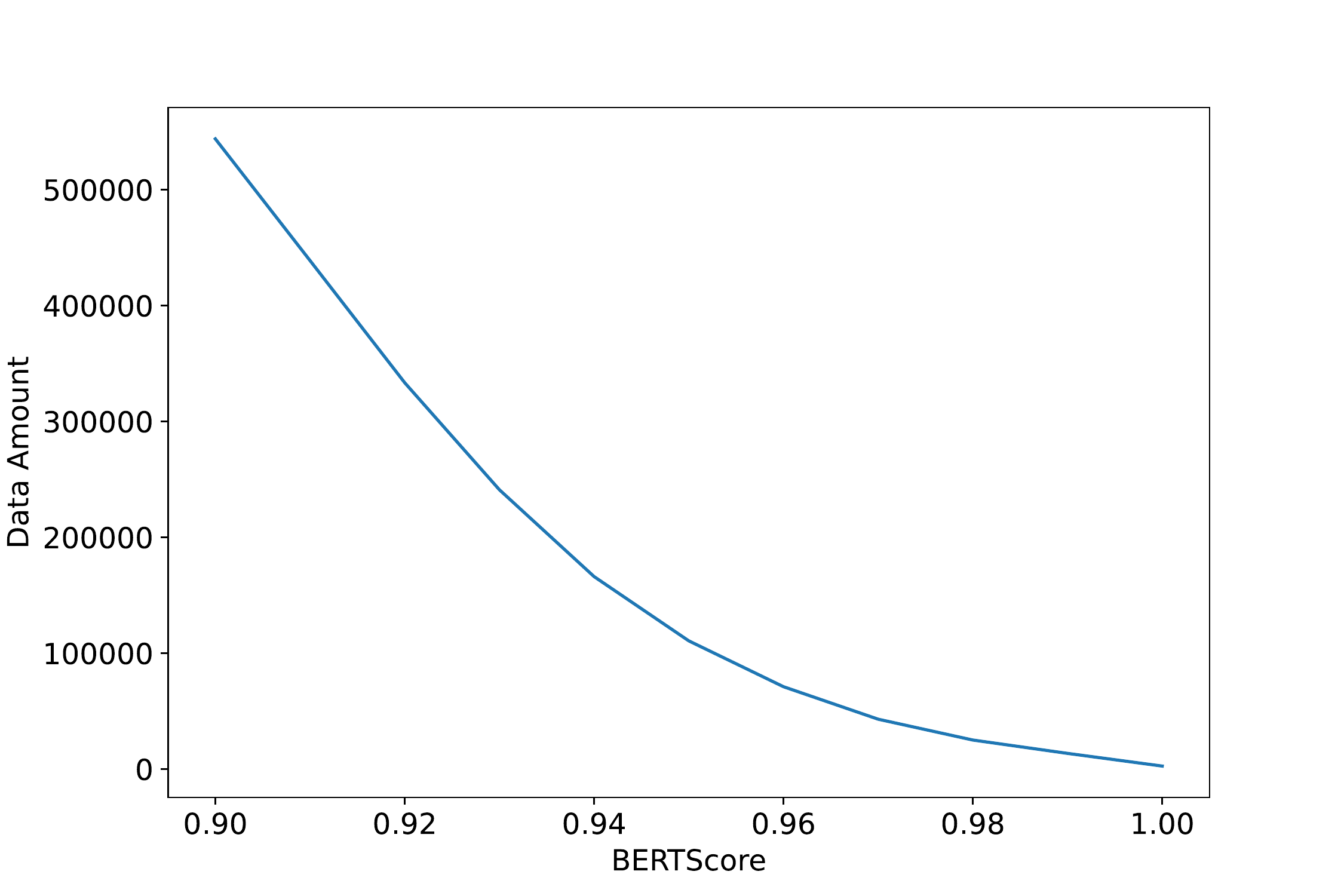}
	\caption{BERTScore with BanglaBERT embeddings for IndicParaphrase Dataset for Range [0.90-1.0]}
	\label{fig:bbert_indic_all_0.9_1.0}
\end{figure}

We see a different scenario for the case of BERTScore (calculated with BanglaBERT embedding) vs. the data amount plot for the whole dataset. In Figure \ref{fig:bbert_indic_all_0.9_1.0}, we observe by taking a closer look at BERTScore for the range of [0.9 - 1.0] that the amount of sentences for threshold of 0.92 is about 0.31M (35\% of the whole dataset) and for 0.93 about 0.23M sentences (about 25\% of the whole dataset). Compared to our dataset, for a threshold above 0.92 for BERTScore, we have a little more than 0.5M (about 37\% of our dataset), and for 0.93, we have about 0.367M sentences (about 27\% of our whole dataset), as seen in Figure \ref{fig:BBERTScore_0.9_1.0}. This indicates that semantic meaning is more preserved in our dataset as we only took the sentences that ensured high semantics in the whole corpus for constructing our final BanglaParaphrase dataset.

\begin{figure}[!tbh]
  \centering
  \includegraphics[width=0.52\textwidth]{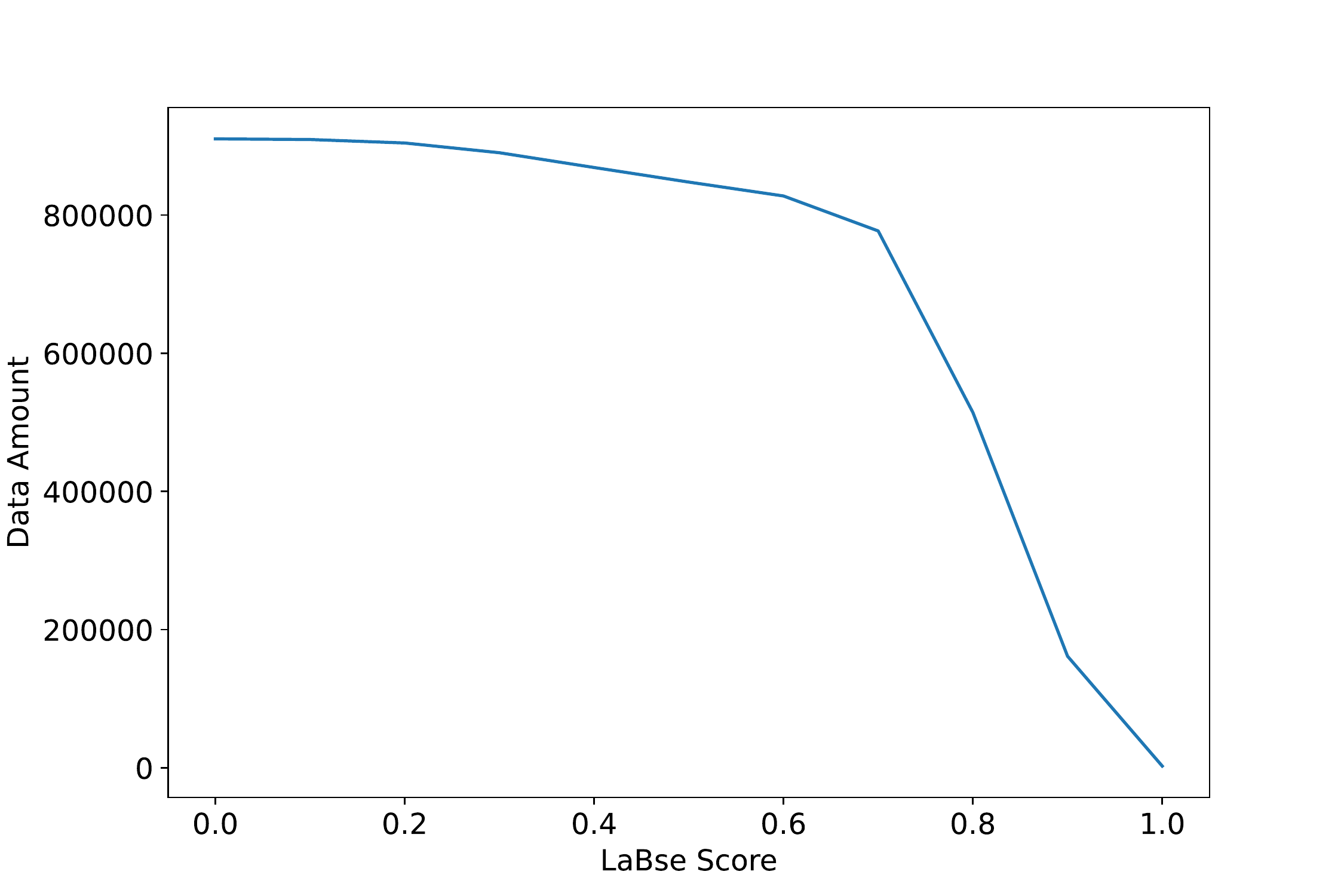}
  \caption{LaBSE Similarity Score for range [0-1.0]}
  \label{fig:labse_indic_all_0_1.0}
\end{figure}

We also observe an analysis with LaBSE similarity score for IndicParaphrase dataset\footnote{only scores above 0 are shown in the plots} where we follow from Figure \ref{fig:labse_indic_all_0_1.0} that above 0.6 there are about more than 0.8M sentences which drastically reduces as the threshold rises. We also observe that above 0.7, there are close to 0.8M sentences. If we look above 0.8, we find that the value drastically reduces to a little more than 0.5M sentences, which is just about 57\% of the total data. If we look above 0.85, we only find about 0.35M sentences, which is about 38\% of the total data available, and it corresponds closely to the amount of 0.31M for BERTScore of 0.92 or above that we discussed.

So the analysis leads us to the inference that the IndicParaphrase dataset is diverse, but it falls short in terms of semantics between the source and the references.